\documentclass{article}
\usepackage{iclr2027_conference,times}

\usepackage{graphicx}
\usepackage{amsmath}
\usepackage{amssymb}
\usepackage{booktabs}
\usepackage{colortbl}
\usepackage{hyperref}
\usepackage{url}
\usepackage{fancyvrb}

\title{Beyond Saying Less: Fine-Grained Alignment for Informative and Faithful Vision-Language Models}

\author{Xingming Long$^{1,2,3}$, Jie Zhang$^{1,2}$\thanks{Corresponding author.}, Yuecong Min$^{1,2}$, Shiguang Shan$^{1,2}$, Xilin Chen$^{1,2}$ \\
$^{1}$State Key Laboratory of AI Safety, Institute of Computing Technology, Chinese Academy of Sciences\\
$^{2}$University of Chinese Academy of Sciences \: $^{3}$Zhongguancun Academy \\
\texttt{xingming.long@vipl.ict.ac.cn,} \\ 
\texttt{\{zhangjie,minyuecong,sgshan,xlchen\}@ict.ac.cn}
}

\iclrfinalcopy

\begin{document}
\maketitle

% —— ICLR: 显示作者，但移除“Published as … / Under review …”横幅 ——
% \usepackage{fancyhdr} % 需要在重定义样式前加载
\makeatletter
% \iclrfinalcopy                 % 1) 开启 camera-ready 行为（显示作者）
% 2) 兼容不同年份的横幅宏：统统置空
\@ifundefined{iclrfinalcopyheader}{}{\renewcommand{\iclrfinalcopyheader}{}}
\@ifundefined{iclrfinalcopyheaderfirstpage}{}{\renewcommand{\iclrfinalcopyheaderfirstpage}{}}
\@ifundefined{iclrfinalcopycomment}{}{\renewcommand{\iclrfinalcopycomment}{}}
% 3) 一些版本把第一页设成专用样式名（如 firstpage / iclrfirstpage / plain）
%    逐个覆盖为“空页眉脚”
\fancypagestyle{firstpage}{\fancyhf{} \renewcommand{\headrulewidth}{0pt}\renewcommand{\footrulewidth}{0pt}}
\fancypagestyle{iclrfirstpage}{\fancyhf{} \renewcommand{\headrulewidth}{0pt}\renewcommand{\footrulewidth}{0pt}}
\fancypagestyle{plain}{\fancyhf{} \renewcommand{\headrulewidth}{0pt}\renewcommand{\footrulewidth}{0pt}}
% 4) 默认页眉脚也清空（如需页码，见下方可选设置）
\pagestyle{fancy}
\fancyhf{}
\renewcommand{\headrulewidth}{0pt}
\renewcommand{\footrulewidth}{0pt}
\makeatother
% —— 可选：如果想保留页码，把这一行打开 —— 
\fancyfoot[C]{\thepage}

\begin{abstract}
Object hallucination remains a major challenge for large vision-language models. While off-policy preference optimization proves to be an effective solution, on-policy reinforcement learning provides a more promising direction as it directly targets a model's current failure modes. However, we find that without fine-grained reward formulation and allocation, on-policy optimization often falls into an easy shortcut: reducing hallucinations merely by saying less---making fewer valid claims.
To comprehensively resolve this, we propose a fine-grained alignment framework that couples dense reward signals at the data level with precise credit assignment at the algorithmic level.
Specifically, we first construct the \textbf{D}ense \textbf{O}bject \textbf{P}resence and \textbf{A}bsence (DOPA) dataset to address sparse annotations that prevent valid object claims from being verified and rewarded. DOPA exhaustively annotates the deterministic presence and absence of every concept across an expanded vocabulary, significantly increasing the density of reliable reward signals during on-policy rollouts.
Second, we propose \textbf{S}ubsentence-level \textbf{C}redit \textbf{A}ssignment for on-\textbf{P}olicy \textbf{O}ptimization (SCAPO) to prevent response-level shared advantages from allowing local hallucinations to compromise all other valid outputs within the same response. By assigning credit to each subsentence independently based on its object claims, SCAPO can precisely reinforce faithful generations and penalize hallucinations.
Furthermore, we leverage the resulting faithful image descriptions as auxiliary context to transfer generative gains to discriminative tasks. Experiments demonstrate that our method produces highly informative, faithful descriptions in generative tasks while yielding clear performance gains on discriminative evaluation.
\end{abstract}

\section{Introduction}

\begin{figure}[t]
    \centering
    \includegraphics[width=0.62\textwidth]{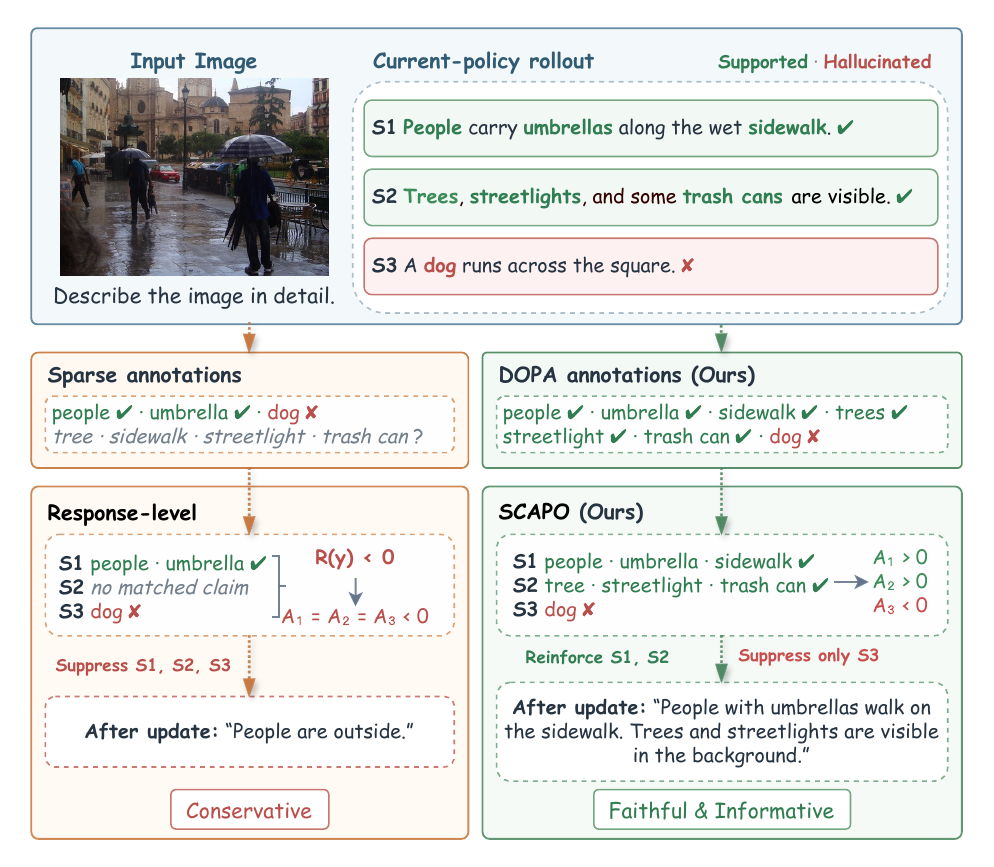}
    \caption{Motivation for our approach. Conventional optimization suffers from sparse annotations leaving valid claims unrewarded, and response-level advantages allowing local hallucinations to suppress correct outputs. By combining the dense rewards of DOPA with the subsentence-level credit assignment of SCAPO, we achieve fine-grained alignment, preventing this suppression and yielding informative and faithful descriptions.}
    \label{fig:motivation}
\end{figure}

Large vision-language models (LVLMs) have achieved substantial progress in image description, visual question answering, and multimodal reasoning. Despite these advances, they frequently generate content that is unsupported by the available visual evidence, a failure commonly referred to as hallucination \cite{LVLM_hal_survey1_2024,LVLM_hal_survey2_2024,LVLM_hal_survey3_2024,LVLM_hal_survey4_2024,POPE_2023}. For example, a model can claim that an object is present even though it does not appear in the image. Such errors reduce response reliability, mislead subsequent reasoning, and impede the deployment of LVLMs in settings that require dependable visual understanding. Ensuring that generated content remains consistently grounded in visual evidence has therefore become a central problem in reliable LVLM research.

Existing hallucination-mitigation approaches broadly fall into inference-time and training-based categories. Inference-time methods operate without retraining by employing contrastive decoding \cite{VCD_2024,ICD_2024}, representation interventions \cite{OPERA_2024,DAMRO_2024,VTI_2025}, or post-hoc self-correction \cite{Woodpecker_2024,LURE_2024}. In contrast, training-based methods intrinsically align LVLM generation preferences. While most existing paradigms rely on off-policy optimization using static datasets \cite{RLHFV_2024,HaDPO_2023,HSA_DPO_2025,POVID_2024,HALVA_2025,DPO_2023}, recent studies demonstrate that learning directly from annotated on-policy rollouts \cite{OPA_DPO_2025} is substantially more effective in mitigating policy-data mismatch.

While optimizing on-policy rollouts directly exposes a model's current errors, we find that without fine-grained reward formulation and allocation, it often falls into an easy shortcut: the model reduces hallucinations not by improving faithfulness, but by reducing informativeness---that is, essentially saying less by making fewer valid claims. As illustrated in Figure~\ref{fig:motivation}, we identify two coarse-grained bottlenecks driving this behavior. The first is a data-level bottleneck: sparse annotations leave valid object claims in current rollouts unverified and unrewarded. The second is an algorithmic-level bottleneck: response-level optimization assigns shared advantages across the entire text, allowing local hallucinations to compromise all other valid outputs within the same response. Consequently, the model is incentivized toward global conservatism rather than precise error correction, underscoring the critical need for denser reward coverage and localized credit assignment.

To comprehensively resolve these bottlenecks, we propose a fine-grained alignment framework coupling dense reward signals with precise credit assignment. At the data level, we construct the Dense Object Presence and Absence (DOPA) dataset, which provides dense, reliable training rewards by exhaustively annotating the deterministic presence or absence of concepts over an expanded vocabulary. At the algorithmic level, we introduce Subsentence-level Credit Assignment for on-Policy Optimization (SCAPO). Rather than relying on coarse response-level shared advantages, SCAPO segments responses at subsentence boundaries and independently assigns credit. This localized mechanism ensures that reinforcement and suppression are precisely targeted, preventing isolated hallucinations from penalizing other valid outputs within the same response.

Our contributions are summarized as follows:
\begin{itemize}
    \item We construct the \textbf{D}ense \textbf{O}bject \textbf{P}resence and \textbf{A}bsence (DOPA) dataset to overcome data-level reward sparsity. By exhaustively annotating the deterministic presence and absence of concepts across an expanded vocabulary, DOPA provides dense and reliable reward signals for object claims during on-policy rollouts.
    \item We propose \textbf{S}ubsentence-level \textbf{C}redit \textbf{A}ssignment for on-\textbf{P}olicy \textbf{O}ptimization (SCAPO) to resolve the algorithmic bottleneck of response-level shared advantages. By assigning credit independently to each subsentence based on its specific claims, SCAPO prevents local hallucinations from suppressing all other valid outputs.
    \item Extensive experiments show that our fine-grained alignment framework overcomes the ``saying less'' shortcut, producing descriptions that are both faithful and informative. As an additional transfer evaluation, we further show that these improved descriptions provide useful auxiliary context for discriminative tasks.
\end{itemize}

\section{Related Work}

\subsection{RLHF and Preference Optimization}
Reinforcement learning from human feedback (RLHF) fundamentally aligns large models using on-policy algorithms such as PPO \cite{PPO_2017}, though maintaining separate reward and value models is computationally expensive. Direct Preference Optimization (DPO) \cite{DPO_2023} provides a simpler offline alternative, while policy-data mismatch has motivated iterative preference-refreshing approaches \cite{IterativeReasoningPO_2024}. More recently, methods such as GRPO \cite{DeepSeekMath_2024} have revitalized on-policy optimization by estimating advantages from relative rewards without requiring a critic model, with subsequent variants addressing issues such as length-biased credit weighting \cite{DAPO_2025,DrGRPO_2025}. However, GRPO-style methods typically assign rewards at the response level. This granularity is particularly limiting for hallucination mitigation, where hallucinated content often constitutes only a small portion of an otherwise faithful response. This motivates finer-grained credit assignment that can selectively penalize hallucinations while reinforcing correct content.

\subsection{Inference-Time Hallucination Mitigation}
Hallucination mitigation in LVLMs has attracted extensive attention \cite{LVLM_hal_survey1_2024,LVLM_hal_survey2_2024,LVLM_hal_survey3_2024,LVLM_hal_survey4_2024}. A prominent line of research focuses on inference-time methods that improve visual faithfulness without updating model parameters. These include contrastive decoding to suppress language priors \cite{VCD_2024,ICD_2024}, generation interventions through attention or logit manipulation \cite{OPERA_2024,DAMRO_2024,VTI_2025}, post-hoc self-correction \cite{Woodpecker_2024,LURE_2024}, external visual-evidence augmentation using auxiliary vision models \cite{MARINE_2025,VEP_2025}, and staged prompting that verbalizes visual evidence before final prediction \cite{VCAP_2025}. However, while inference-time approaches can improve visual faithfulness without retraining, they inevitably introduce inference latency and fail to internalize visual faithfulness directly into the model parameters.

\subsection{Training-Based Hallucination Mitigation}
Training-based methods intrinsically improve visual grounding via preference optimization. Early offline approaches construct preference pairs using expert corrections \cite{RLHFV_2024,HaDPO_2023,HSA_DPO_2025} or synthetic hallucination injection \cite{POVID_2024,HALVA_2025}. However, recent studies \cite{RLAIFV_2025,OPA_DPO_2025} demonstrate that on-policy training is more effective, as it directly targets the evolving policy's failure modes. Yet, existing online methods face critical bottlenecks: accurate credit assignment demands either complex rollout modification \cite{OPA_DPO_2025} or computationally expensive iterative LLM judge evaluation \cite{RLAIFV_2025}. To overcome these challenges, we introduce the DOPA dataset as a reusable, high-density reward oracle providing exhaustive object-existence annotations on an expanded vocabulary. Building on this, we propose SCAPO to leverage these dense signals, shifting from coarse response-level evaluation to precise subsentence-level credit assignment for on-policy rollouts.

\begin{figure}[t]
    \centering
    \includegraphics[width=\textwidth]{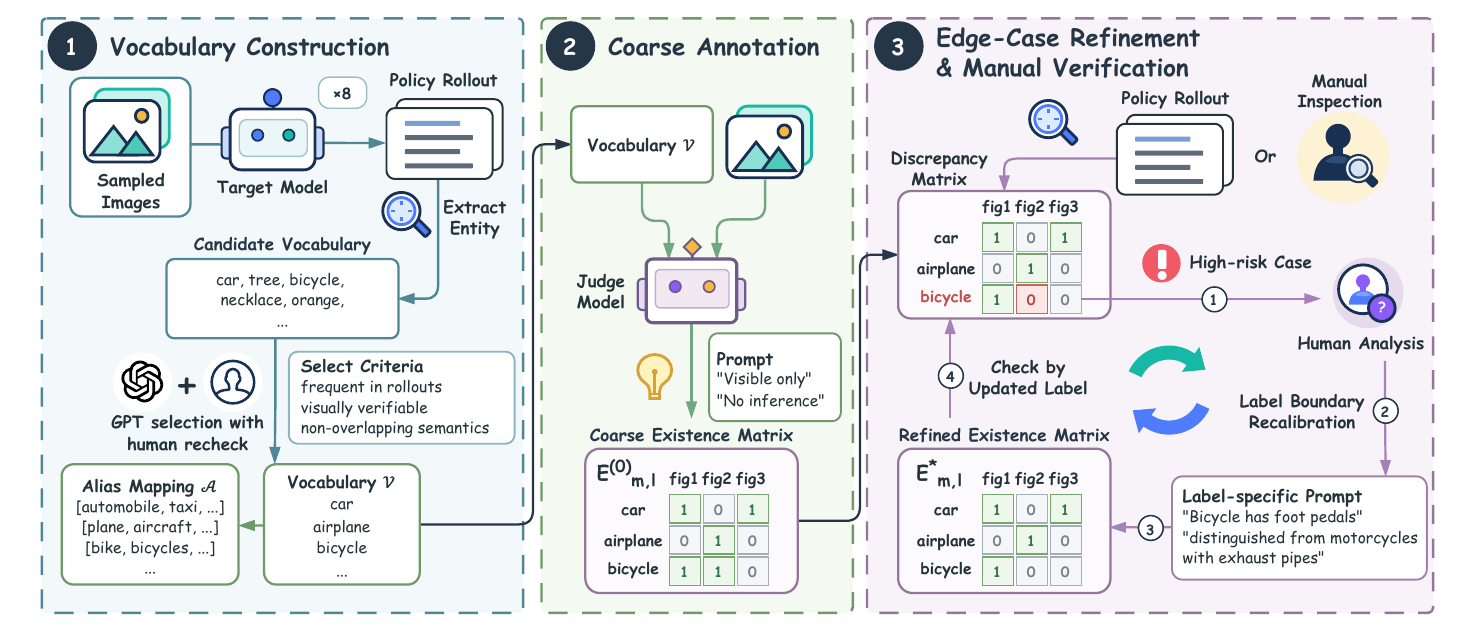}
    \caption{Pipeline for constructing the DOPA dataset. The process begins with rollout-driven vocabulary construction to establish the canonical vocabulary $\mathcal{V}$ and alias mapping $\mathcal{A}$. Based on these, large-scale coarse annotation and edge-case refinement are performed. Ultimately, a rigorous manual verification cycle yields the high-fidelity object-existence matrix $E^\ast$.}
    \label{fig:dataset_pipeline}
\end{figure}

\section{Construction of DOPA Dataset}

To overcome the data-level bottleneck where sparse annotations leave valid object claims unrewarded, we construct the Dense Object Presence and Absence (DOPA) dataset to provide dense and reliable reward signals for on-policy rollouts. We build DOPA on 5,000 images sampled from the MS COCO \cite{MSCOCO_2014} training split, disjoint from all evaluation images. As illustrated in Figure~\ref{fig:dataset_pipeline}, its construction comprises vocabulary construction, coarse annotation, edge-case refinement, and manual verification, which are detailed in the following subsections.

\subsection{Vocabulary Construction}

To increase reward coverage for concepts naturally mentioned by the target model, we derive an expanded vocabulary directly from its rollouts on the training data. After extracting and normalizing entity mentions, we curate these candidates into canonical object labels based on three strict criteria: \textbf{(i)} sufficient frequency in target-model rollouts; \textbf{(ii)} direct visual verifiability without subjective inference; and \textbf{(iii)} semantic clarity with minimal overlap. We initially use GPT-5.5 \cite{GPT55_2026} to cluster the candidates into a draft set, and then manually refine this draft against the aforementioned criteria. Finally, we construct a conflict-free alias mapping covering synonyms and morphological variations for each retained label. This process yields the final vocabulary $\mathcal{V}$ comprising \textbf{160} canonical labels and a mapping $\mathcal{A}$ of \textbf{894} aliases.

\subsection{Coarse Annotation}

Existing datasets typically provide only positive annotations, leaving it ambiguous whether an unannotated concept is absent or simply omitted. DOPA instead defines a closed-set object-label space over \(\mathcal V\), in which every image-label pair is explicitly annotated as either present or absent. This exhaustive presence-absence annotation removes the unknown state for all in-vocabulary claims and provides unambiguous reward signals during on-policy training.
We employ Qwen3.6-27B \cite{Qwen36_27B_2026} as the initial coarse judge to implement this exhaustive annotation. Prompted with strict criteria for handling occlusion and explicitly prohibiting commonsense inference, the judge determines object presence based solely on visual evidence. This produces a complete initial object-existence matrix $E^{(0)}_{m,l}\in\{0,1\}$, indicating whether the object corresponding to the $m$-th canonical label is present in the $l$-th image.

\subsection{Edge-Case Refinement \& Manual Verification}

To improve annotation fidelity, we implement an iterative refinement and manual verification protocol. First, we evaluate model-generated object claims against the coarse matrix to isolate high-risk edge cases, such as heavy occlusion or unclear semantic boundaries. We resolve these systematic ambiguities by crafting tailored, label-specific prompts with rigorous inclusion/exclusion criteria to re-query the judge model. Subsequently, for each canonical label, we randomly sample 30 image-label pairs for manual evaluation. After each round of prompt refinement and re-annotation, we draw a new random set of 30 pairs and repeat the evaluation until no errors are observed in the newly sampled subset. Once every label passes this sample-based quality-control procedure, we finalize the re-annotated object-existence matrix $E^\ast_{m,l}\in{0,1}$.

Ultimately, this exhaustive process provides dense and reliable reward signals for the subsentence-level credit assignment, enabling DOPA to serve as a high-quality, reusable reward oracle during training that requires only deterministic lookups rather than repeated judge-model calls.

\section{SCAPO Method}

\begin{figure}[t]
    \centering
    \includegraphics[width=\textwidth]{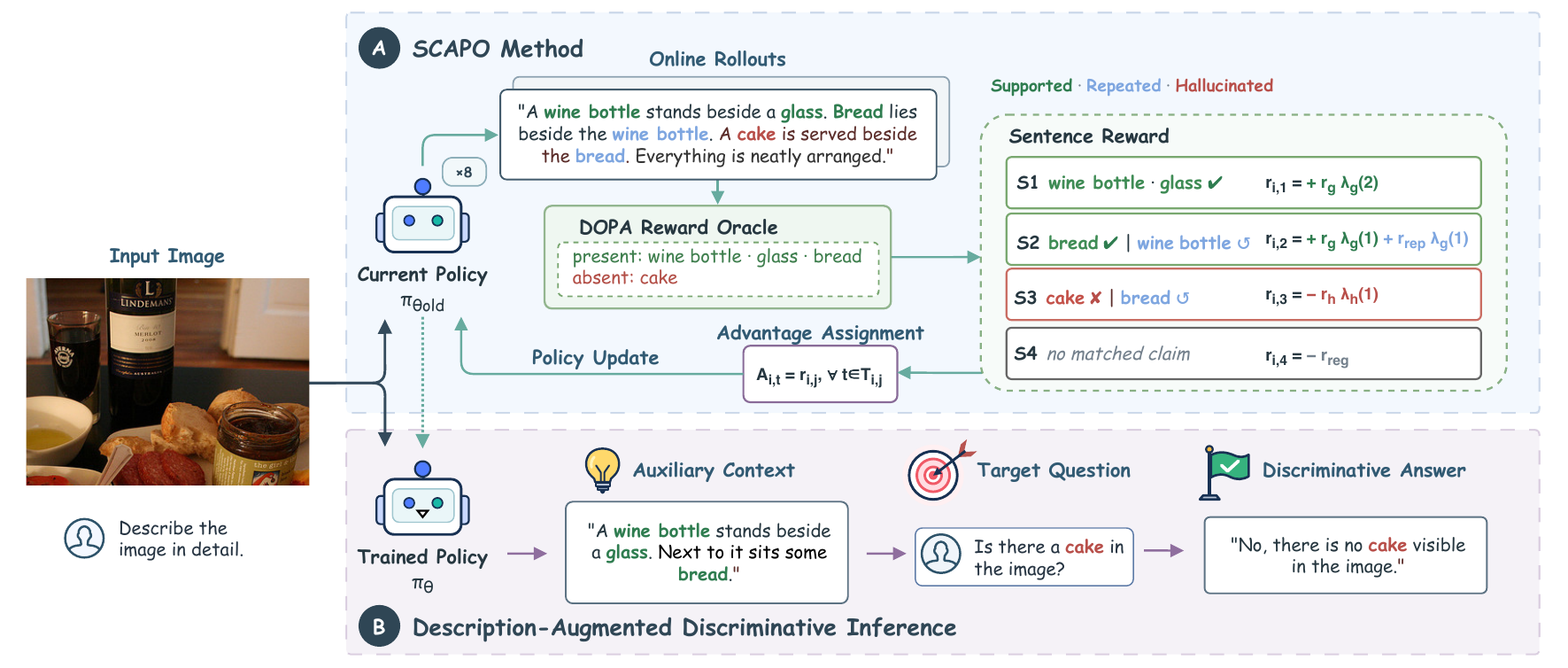}
    \caption{Overview of the SCAPO method and the adopted inference paradigm. Current-policy rollouts are segmented at subsentence boundaries and evaluated against the DOPA reward oracle. The resulting subsentence-level rewards are independently assigned to their corresponding token spans for the on-policy update. Furthermore, the optimized model leverages its generated faithful descriptions as auxiliary context for description-augmented discriminative inference.}
    \label{fig:method_overview}
\end{figure}

To prevent coarse response-level shared advantages from allowing local hallucinations to suppress valid outputs, we build on the DOPA dataset and propose Subsentence-level Credit Assignment for on-Policy Optimization (SCAPO). As illustrated in Figure~\ref{fig:method_overview}, SCAPO segments each response at subsentence boundaries and evaluates the object claims within each subsentence against the DOPA annotations, effectively encouraging highly informative and faithful generations rather than leading to global conservatism. We provide formal definitions and the pseudocode of our SCAPO method in Appendix~\ref{app:formal}.

\subsection{Subsentence-Level Reward}

To evaluate the object mentions within the $j$-th subsentence of the $i$-th response, we categorize them according to the DOPA annotations and define three explicit counts: $n^{\mathrm{new}}_{i,j}$ represents the number of newly covered supported objects (valid claims appearing for the first time in the response), $n^{\mathrm{rep}}_{i,j}$ denotes the repeated supported objects (valid claims already mentioned in preceding subsentences), and $n^{\mathrm{hal}}_{i,j}$ counts the hallucinated objects. The subsentence-level reward is then defined as:
\begin{equation}\label{eq:subsentence_reward}
r_{i,j}
=
\begin{cases}
-r_h\,\lambda_h\!\left(n^{\mathrm{hal}}_{i,j}\right),
& n^{\mathrm{hal}}_{i,j}>0,\\[2mm]
-r_{\mathrm{reg}},
& \text{no objects},\\[2mm]
+r_g\,\lambda_g\!\left(n^{\mathrm{new}}_{i,j}\right)
+r_{\mathrm{rep}}\,\lambda_g\!\left(n^{\mathrm{rep}}_{i,j}\right),
& \text{otherwise}.
\end{cases}
\end{equation}
Here, $r_h, r_g, r_{\mathrm{rep}}, r_{\mathrm{reg}} \geq 0$ are base coefficients, scaled by monotonically increasing functions $\lambda_h$ and $\lambda_g$. This formulation explicitly enforces three design principles:
\textbf{(1) Zero-tolerance for hallucinations:} any hallucination triggers a penalty, strictly preventing correct information from offsetting explicit errors.
\textbf{(2) Grounded informativeness:} hallucination-free subsentences are rewarded for novel claims and, marginally, for repetitions.
\textbf{(3) Anti-conservatism:} subsentences lacking verifiable objects incur a regularization penalty ($-r_{\mathrm{reg}}$) to discourage the model from generating generic, uninformative content.

\subsection{Subsentence-Level On-Policy RL Training}

During each optimization step, we sample image-prompt pairs from the current policy snapshot $\pi_{\theta_{\mathrm{old}}}$. For each response $y_i$, we compute subsentence-level rewards $r_{i,j}$ via Eq.~\eqref{eq:subsentence_reward}. To bypass coarse response-level aggregation, we directly broadcast these local rewards as token advantages: $A_{i,t}=r_{i,j}$ for all tokens $t\in T_{i,j}$, where $T_{i,j}$ is the token span of subsentence $s_{i,j}$. Using these token advantages, we optimize the standard clipped surrogate objective:
\begin{equation}
\mathcal{L}_{\mathrm{SCAPO}}
=
-\frac{1}{N_{\mathrm{tok}}}
\sum_{i,j}\sum_{t\in T_{i,j}}
\min\!\Big(
\rho_{i,t}A_{i,t},
\operatorname{clip}(\rho_{i,t},1-\epsilon,1+\epsilon)A_{i,t}
\Big)
+\beta_{\mathrm{KL}}D_{\mathrm{KL}}\!\left(\pi_\theta\Vert\pi_{\mathrm{ref}}\right),
\end{equation}
where $N_{\mathrm{tok}}$ is the total number of response tokens in the batch, $\rho_{i,t}$ is the standard policy importance ratio, $\epsilon$ is the clipping coefficient, and $\beta_{\mathrm{KL}}$ controls the KL regularization toward the fixed reference policy $\pi_{\mathrm{ref}}$. Thus, while policy optimization remains token-wise, credit is localized at the subsentence level, with KL regularization preventing excessive policy deviation.

Unlike conventional PPO-based RLHF, SCAPO does not require training or maintaining an auxiliary value model. PPO-based RLHF typically estimates token-level advantages through a learned critic and GAE, whereas SCAPO assigns each subsentence an independently verified factual reward that is shared only among its own tokens. Advantage signals are not propagated across subsentence boundaries, preventing a later hallucination from altering the credit assigned to an earlier faithful statement. This design replaces general-purpose return estimation with explicit semantic locality, which is well suited to sparse and localized hallucination errors.

\subsection{Description-Augmented Discriminative Inference}

Following SCAPO optimization, the policy exhibits a marked improvement in generating faithful and informative visual descriptions. To transfer these generative grounding capabilities to downstream discriminative tasks, we adopt the description-augmented inference paradigm. Specifically, given an image and a discriminative query, the optimized model first generates a grounded description. The model then conditions its final discriminative answer on both the original image and this generated description jointly.

\section{Experiments}

\subsection{Experimental Setup}

\paragraph{Datasets.}
We evaluate our approach across both generative and discriminative tasks. For generative tasks, we assess hallucination rate and informativeness using MS COCO \cite{MSCOCO_2014} and AMBER \cite{AMBER_2023}. For discriminative tasks, we utilize PhD \cite{PhD_2025} and the discriminative tasks of AMBER.

\begin{table}[t]
\centering
\caption{Evaluation on generative and discriminative hallucination benchmarks. \textbf{Cap. Score} = \textbf{HarMean(}1-Hal.\ Rate, Cover Rate\textbf{)} jointly measures caption faithfulness and informativeness. (Cap. Score is computed with the original object-existence annotation of the dataset. Cap. Score\(^{\dagger}\) is computed with our expanded DOPA annotation.)}
\label{tab:rule_based_main}
\footnotesize
\setlength{\tabcolsep}{3.8pt}
\resizebox{\textwidth}{!}{%
\begin{tabular}{lccccccccc}
\toprule
& \multicolumn{2}{c}{AMBER} & \multicolumn{2}{c}{MS COCO} & \multicolumn{5}{c}{PhD} \\
\cmidrule(lr){2-3}\cmidrule(lr){4-5}\cmidrule(lr){6-10}
Method
& Cap. Score $\uparrow$ & Dis. Acc. $\uparrow$ & Cap. Score $\uparrow$ & Cap. Score\(^{\dagger}\) $\uparrow$
& PhD-base $\uparrow$ & PhD-sec $\uparrow$ & PhD-icc $\uparrow$ & PhD-ccs $\uparrow$ & PhD-all $\uparrow$ \\
\midrule
\rowcolor[gray]{0.93}
\multicolumn{10}{l}{\textit{(i) Inference-time Methods}} \\
Qwen2.5-VL & 76.5 & 83.9 & 73.2 & 64.2 & 76.8 & 68.2 & 62.1 & 59.5 & 66.6 \\
VCD & 76.5 & 83.8 & 72.7 & 65.9 & 77.0 & \underline{71.1} & 63.6 & 65.2 & 69.2 \\
OPERA & 76.3 & 84.5 & 73.7 & 65.3 & 76.4 & 70.9 & 64.6 & 62.0 & 68.5 \\
VTI & 47.7 & \underline{85.9} & 50.2 & 33.6 & 79.0 & 57.3 & 58.3 & 66.8 & 65.4 \\
\midrule
\rowcolor[gray]{0.93}
\multicolumn{10}{l}{\textit{(ii) Off-policy Methods}} \\
DPO & \underline{77.7} & 71.8 & \underline{74.8} & \underline{70.0} & 70.8 & 52.9 & 52.4 & 67.9 & 61.0 \\
mDPO & 76.3 & 82.9 & 74.2 & 66.5 & 77.0 & 70.0 & \underline{66.7} & 67.4 & \underline{70.3} \\
SymMPO & 75.5 & 85.4 & 74.2 & 65.4 & \underline{80.2} & 62.8 & 55.6 & \textbf{73.3} & 68.0 \\
OPA-DPO & 67.8 & 85.4 & 72.2 & 53.9 & 76.1 & 70.9 & 52.7 & \underline{69.5} & 67.3 \\
\midrule
\rowcolor[gray]{0.93}
\multicolumn{10}{l}{\textit{(iii) On-policy Methods}} \\
GRPO & 69.9 & 83.8 & 66.5 & 51.4 & 77.2 & 63.9 & 63.5 & 61.8 & 66.6 \\
DAPO & 61.1 & 83.1 & 51.8 & 33.3 & 77.1 & 57.9 & 59.8 & 63.3 & 64.5 \\
Dr.\ GRPO & 68.3 & 84.1 & 65.1 & 47.4 & 76.8 & 56.2 & 62.5 & 62.7 & 64.5 \\
\textbf{SCAPO (Ours)} & \textbf{82.8} & \textbf{87.4} & \textbf{79.6} & \textbf{78.7} & \textbf{80.3} & \textbf{74.5} & \textbf{67.5} & 68.2 & \textbf{72.6} \\
\bottomrule
\end{tabular}
}
\end{table}

\paragraph{Comparison Methods.}
All methods use Qwen2.5-VL-7B \cite{Qwen25VL_2025} as the backbone. We organize the evaluated methods into three groups. The inference-time group includes the unmodified backbone, VCD \cite{VCD_2024}, OPERA \cite{OPERA_2024}, and VTI \cite{VTI_2025}. The off-policy group includes standard DPO \cite{DPO_2023}, multimodal DPO (mDPO) \cite{mDPO_2024}, SymMPO \cite{SymMPO_2025}, and OPA-DPO \cite{OPA_DPO_2025}. The preference data for these methods are constructed from Qwen2.5-VL-7B rollouts and annotated by Qwen3.6-27B \cite{Qwen36_27B_2026}. The on-policy group includes GRPO \cite{DeepSeekMath_2024}, DAPO \cite{DAPO_2025}, Dr.\ GRPO \cite{DrGRPO_2025}, and our SCAPO. All the on-policy methods are trained using our DOPA annotations. Complete training configurations are reported in Appendix~\ref{app:train_config}.

\paragraph{Evaluation Metrics.}
For discriminative evaluation, we report standard metrics including accuracy, precision, recall, and F1, as well as the benchmark-specific PhD-index.
For generative evaluation, to enable the metric to simultaneously reflect the faithfulness and informativeness of the model output, we introduce a new metric \textbf{Caption Score (Cap. Score)} built on the harmonic mean of \textbf{1-Hal.\ Rate} and \textbf{Cover Rate}, where Hal.\ Rate corresponds to CHAIR$_i$. The formal definitions are given in Appendix~\ref{app:metrics}.

\subsection{Main Results}

Table~\ref{tab:rule_based_main} presents the main results across three hallucination benchmarks. When simultaneously considering faithfulness and informativeness, most hallucination mitigation methods struggle to yield visible improvements in the Cap. Score metric. Aside from our SCAPO, only DPO achieves consistent gains across the three generative settings. Notably, despite being trained on the same DOPA data as SCAPO, other GRPO-style on-policy methods yield Cap. Scores below the Qwen2.5-VL baseline. This degradation stems from an optimization shortcut toward global conservatism, which we analyze in detail in Section~\ref{sec:exp_scapo}. In contrast, SCAPO effectively overcomes this bottleneck, achieving strong Cap. Scores of 82.8\% on AMBER and 79.6\% on MS COCO.

Importantly, these generative improvements also transfer to discriminative tasks. SCAPO achieves an 87.4\% discriminative accuracy on AMBER and the best results among the compared methods on PhD-base, PhD-sec, PhD-icc, and PhD-all. Overall, this consistent improvement across both generative and discriminative settings distinguishes SCAPO from alternatives that inflate an isolated metric at the expense of cross-benchmark performance.

\subsection{Experiments on DOPA}

\begin{table}[t]
\centering
\caption{Effect of the DOPA dataset used for on-policy training. Both variants below utilize the same SCAPO objective but derive rewards from different annotations.}
\label{tab:dopa_ablation}
\small
\setlength{\tabcolsep}{5.0pt}
\resizebox{\textwidth}{!}{%
\begin{tabular}{l cccccccc}
\toprule
& \multicolumn{2}{c}{MS COCO Label} & \multicolumn{2}{c}{DOPA Label} & \multicolumn{2}{c}{AMBER} & \multicolumn{2}{c}{MMHal-Bench} \\
\cmidrule(lr){2-3}\cmidrule(lr){4-5}\cmidrule(lr){6-7}\cmidrule(lr){8-9}
Method
& Hal.\ Rate $\downarrow$ & Cover Rate $\uparrow$
& Hal.\ Rate $\downarrow$ & Cover Rate $\uparrow$
& Hal.\ Rate $\downarrow$ & Cover Rate $\uparrow$
& Score $\uparrow$ & Hal.\ Rate $\downarrow$ \\
\midrule
Qwen2.5-VL (Baseline) & 15.1 & 64.3 & 11.8 & 50.5 & 4.9 & 64.0 & 3.26 & 39.6 \\
\midrule
\rowcolor[gray]{0.93}
\multicolumn{9}{l}{\textit{SCAPO Training}} \\
\quad w/ MS COCO annotations & 13.9 & \textbf{78.8} & 11.6 & 52.0 & \textbf{4.6} & 66.1 & 3.45 & 34.4 \\
\quad w/ DOPA annotations (\textbf{Ours}) & \textbf{13.4} & 73.7 & \textbf{9.4} & \textbf{69.6} & 4.8 & \textbf{73.2} & \textbf{3.52} & \textbf{33.3} \\
\bottomrule
\end{tabular}
}
\end{table}

\paragraph{Generalization via DOPA Annotations.} 
Table~\ref{tab:dopa_ablation} isolates the effect of the annotation source while keeping the SCAPO objective unchanged. We additionally incorporate the LLM-judge-based MMHal-Bench \cite{LLaVARLHF_2023} for rigorous cross-benchmark assessment. It can be observed that while training with sparse MS COCO annotations leads to substantial coverage improvements on its specific label set, the gains on other benchmarks are notably smaller, as claims outside this space remain unverifiable and unrewarded. In contrast, replacing these sparse labels with the dense DOPA dataset drives generalized improvements. It significantly boosts in-domain performance under the DOPA evaluation, reducing the Hal.\ Rate from 11.8\% to 9.4\% and raising the Cover Rate from 50.5\% to 69.6\%. Furthermore, it exhibits superior cross-dataset generalization, maintaining an overall advantage over the sparsely annotated counterpart on AMBER and MMHal-Bench.

\begin{table}[t]
\centering
\caption{Annotation density on training-set rollouts. For each annotation setting, we sample eight responses per training example and report the per-response averages. ``Sup.'' and ``Hal.'' denote ground-truth-supported and hallucinated content, respectively.}
\label{tab:dopa_reward_density}
\small
\setlength{\tabcolsep}{5.0pt}
\resizebox{0.7\textwidth}{!}{%
\begin{tabular}{l *{6}{>{\centering\arraybackslash}p{1.1cm}}}
\toprule
& \multicolumn{3}{c}{Annotated Objects per Response}
& \multicolumn{3}{c}{Subsentences per Response} \\
\cmidrule(lr){2-4}\cmidrule(lr){5-7}
Annotation
& Avg. & Sup. & Hal.
& Avg. & Sup. & Hal. \\
\midrule
MS COCO & 2.73 & 2.32 & 0.41 & 6.33 & 2.62 & 0.45 \\
DOPA (\textbf{Ours}) & 5.13 & 4.37 & 0.76 & 6.37 & 3.46 & 0.81 \\
\bottomrule
\end{tabular}
}
\end{table}

\paragraph{Reward Density Analysis.}
The gains from DOPA can be largely explained by its higher reward density, as demonstrated in Table~\ref{tab:dopa_reward_density}. Compared to MS COCO annotations, DOPA nearly doubles the average number of annotated objects (5.13 vs. 2.73) and significantly increases the total number of reward-bearing subsentence labels. This richer provision of verifiable feedback effectively grounds the on-policy optimization, preventing valid object claims from going unrewarded.

\subsection{Analysis of SCAPO}
\label{sec:exp_scapo}

\begin{figure}[t]
    \centering
    \includegraphics[width=0.65\textwidth]{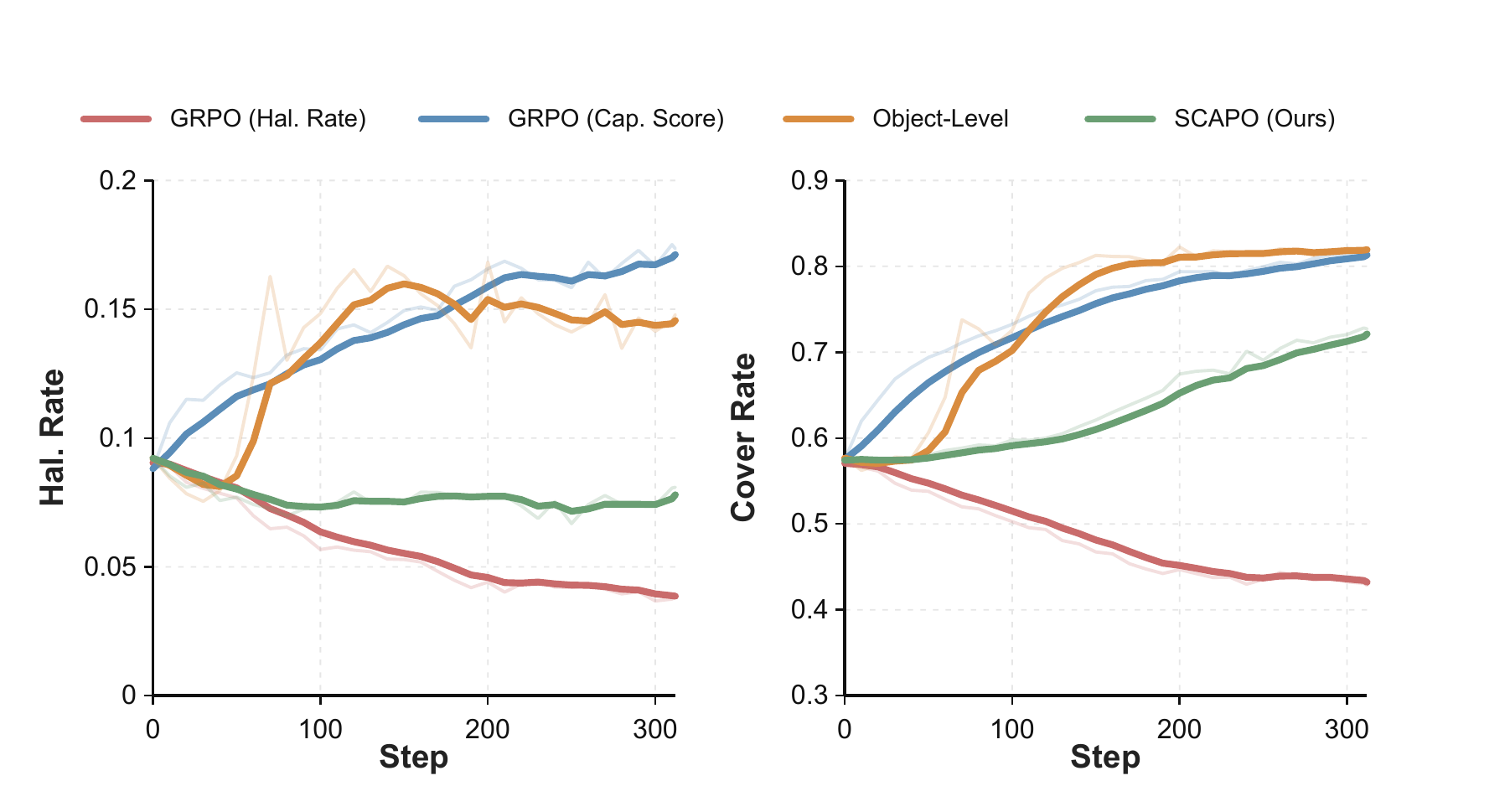}
    \caption{Training dynamics under different credit-assignment strategies. Response-level objectives face a trade-off between limited coverage (when optimizing Hal.\ Rate) and increased hallucinations (when optimizing Cap.\ Score). Object-level assignment fragments semantic coherence, causing unstable increases in both metrics. In contrast, SCAPO's subsentence-level credit assignment resolves these issues, steadily expanding valid coverage while keeping the hallucination rate controlled.}
    \label{fig:train_curve}
\end{figure}

\paragraph{Training Dynamics.}
Figure~\ref{fig:train_curve} compares SCAPO against alternative credit assignment strategies to illustrate their distinct failure modes. Optimizing solely for a response-level Hal.\ Rate penalty triggers the conservative ``saying less'' shortcut, causing Cover Rate to plummet despite dropping Hal.\ Rate to 4\%. Conversely, optimizing for a response-level Cap.\ Score fails to effectively suppress hallucinations, raising Cover Rate but driving Hal.\ Rate up to 17\%. Furthermore, an object-level variant fragments semantic coherence, yielding a highly unstable curve where both metrics increase. SCAPO overcomes these pitfalls through subsentence-level credit assignment. By evaluating claims within their natural semantic boundaries, SCAPO steadily improves Cover Rate while keeping Hal.\ Rate firmly controlled, successfully avoiding both global conservatism and excessive hallucinations.

\begin{figure}[t]
    \centering
    \includegraphics[width=0.65\textwidth]{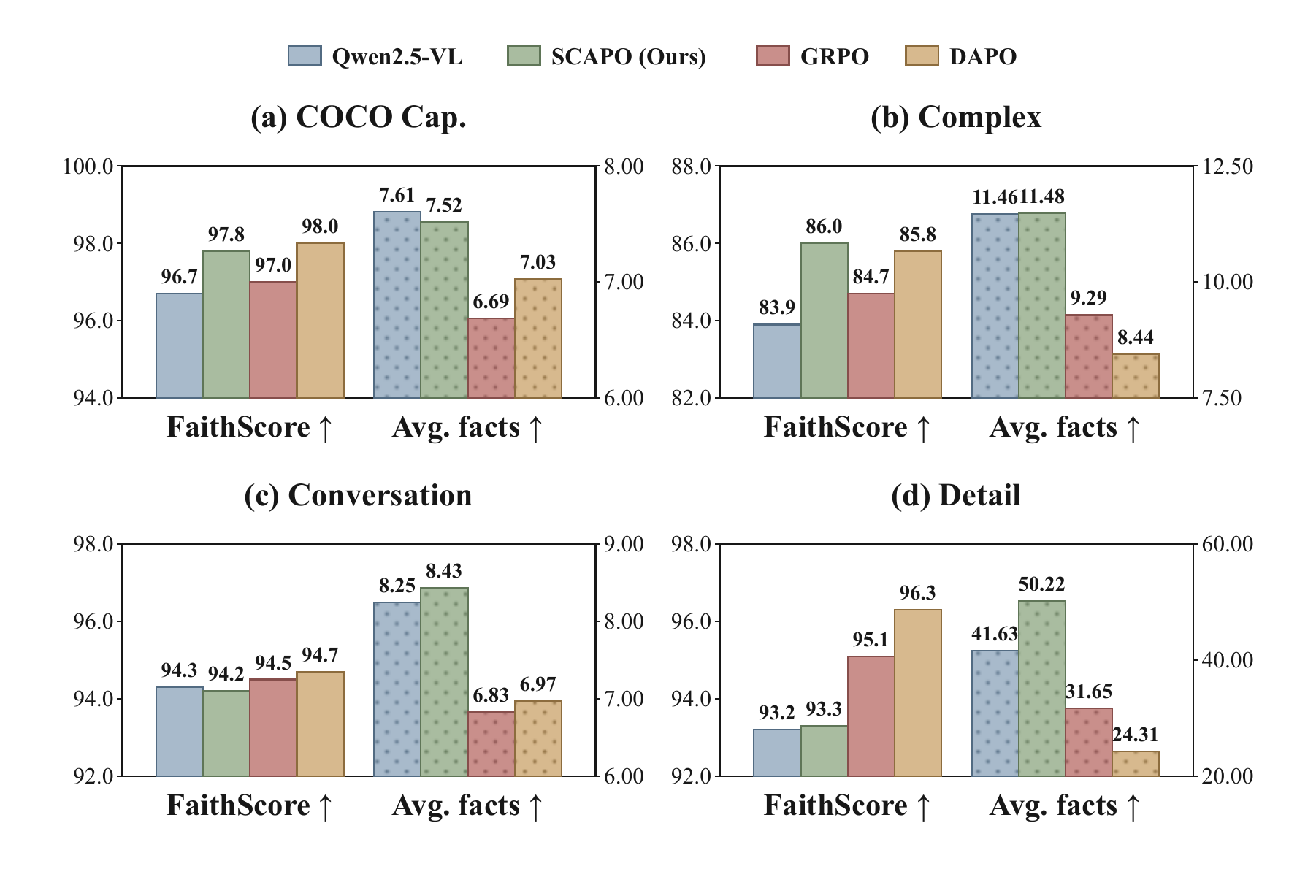}
    \caption{FaithScore and Avg. facts across four generation settings. GRPO-style methods all improve faithfulness at the expense of the number of atomic facts in the output, while SCAPO preserves factual richness and maintains high faithfulness.}
    \label{fig:faithscore_analysis}
\end{figure}

\paragraph{Output Factual Richness.}
Because Cover Rate is bounded by a predefined vocabulary, we further evaluate open-vocabulary performance using the FaithScore benchmark \cite{FaithScore_2024}. This benchmark employs an LLM judge to decompose responses into atomic facts and reports two key metrics: the FaithScore metric (which measures overall faithfulness, where higher indicates fewer hallucinations) and Avg.\ Facts (which measures informativeness by counting the average number of atomic facts per response). As illustrated in Figure~\ref{fig:faithscore_analysis}, response-level GRPO-style methods achieve high FaithScores by exploiting a conservative shortcut: they severely curtail their outputs across all four evaluation settings. For instance, under the detailed setting, the Avg.\ Facts for GRPO and DAPO drop to 31.65 and 24.31, respectively, compared to 41.63 for the Qwen2.5-VL backbone. In contrast, SCAPO secures a superior FaithScore while simultaneously preserving the Avg.\ Facts of the backbone across all settings. This provides compelling evidence that SCAPO's improvements stem from precise hallucination suppression rather than global output impoverishment.

\subsection{Description-Augmented Discriminative Inference}

\begin{table}[t]
\centering
\caption{Progressive ablation on the AMBER discriminative task. We first augment the Qwen2.5-VL baseline with a self-generated image description, and then upgrade the underlying backbone to the SCAPO-trained policy to evaluate the impact of generation quality.}
\label{tab:description_augmented}
\small
\setlength{\tabcolsep}{5.0pt}
\resizebox{0.6\columnwidth}{!}{%
\begin{tabular}{lcccc}
\toprule
Configuration & Acc. $\uparrow$ & Pre. $\uparrow$ & Rec. $\uparrow$ & F1 $\uparrow$ \\
\midrule
Qwen2.5-VL & 83.9 & 84.5 & 92.6 & 88.4 \\
$+$ Desc. Aug.
& 86.9{\scriptsize\,\textcolor{red}{(+3.0)}}
& 87.3{\scriptsize\,\textcolor{red}{(+2.8)}}
& 93.8{\scriptsize\,\textcolor{red}{(+1.2)}}
& 90.4{\scriptsize\,\textcolor{red}{(+2.0)}} \\
$+$ SCAPO
& \textbf{87.4}{\scriptsize\,\textcolor{red}{(+3.5)}}
& \textbf{87.6}{\scriptsize\,\textcolor{red}{(+3.1)}}
& \textbf{94.4}{\scriptsize\,\textcolor{red}{(+1.8)}}
& \textbf{90.9}{\scriptsize\,\textcolor{red}{(+2.5)}} \\
\bottomrule
\end{tabular}
}
\end{table}

Table~\ref{tab:description_augmented} presents a progressive ablation of the description-augmented discriminative inference. First, adding a generated description to the direct-answering baseline improves the Qwen2.5-VL backbone's accuracy from 83.9\% to 86.9\% and its F1 score from 88.4\% to 90.4\%, confirming the general utility of an explicit auxiliary visual representation. Next, substituting the generative backbone with our SCAPO-aligned model further elevates accuracy to 87.4\% and F1 to 90.9\%. These progressive gains indicate that the effectiveness of description augmentation depends on the quality of the generated visual context. By supplying denser valid evidence and strictly fewer misleading priors, SCAPO transfers its generative grounding improvements to downstream discrimination, requiring zero task-specific fine-tuning.

\section{Conclusion}

In this paper, we investigate the conservative ``saying less'' shortcut that plagues on-policy hallucination mitigation, attributing this degradation to sparse reward supervision and coarse response-level credit assignment. To comprehensively resolve these bottlenecks, we introduce a dual-level fine-grained alignment framework. At the data level, we construct the DOPA dataset as a dense and reliable reward oracle; at the algorithmic level, we propose SCAPO for precise subsentence-level credit assignment. Extensive experiments across multiple hallucination benchmarks demonstrate that our approach effectively resolves the faithful-informative trade-off, substantially reducing unsupported content while expanding the coverage of valid object claims. Furthermore, we show that these faithful and informative descriptions serve as highly effective auxiliary context, successfully transferring our generative gains to downstream discriminative tasks.

\section*{AI Use Statement}

In this work, we used generative AI tools for dataset construction, which we disclose as part of the required AI-use disclosure. Specifically, GPT-5.5 \citep{GPT55_2026} was used to draft a canonical object vocabulary, and Qwen3.6-27B \citep{Qwen36_27B_2026} was used to generate the initial and refined object-existence annotations for DOPA. We did not use generative AI to generate research hypotheses, design the proposed method, or derive scientific conclusions. In addition, we used generative AI tools for recommended-disclosure tasks, including language polishing of author-written text and assistance with routine coding. All AI-assisted text was reviewed for consistency with the authors' intended meaning, and AI-assisted code was inspected and tested by the authors. We take responsibility for the final content of this work, including all text, claims, code, and artifacts produced with the aid of generative AI.

\section*{Reproducibility Statement}

We take several steps to facilitate reproducibility. The appendix provides the complete formalization of our reward construction and optimization procedure, including the claim-set definitions and pseudocode for SCAPO (Appendix~\ref{app:formal}). We report the full experimental configuration, including training hyperparameters and environment, in Appendix~\ref{app:train_config}. All prompts used for data annotation, training, and inference are provided verbatim in Appendix~\ref{app:prompts}.

\bibliography{references}
\bibliographystyle{iclr2027_conference}

\providecommand{\PH}[1]{\textcolor{red}{\textbf{[TODO: #1]}}}
\providecommand{\PHBLOCK}[1]{\par\noindent\textcolor{red}{\textbf{[TODO --- #1]}}\par}

% Helpers for the hand-set pseudocode in Figure \ref{alg:scapo}.  Defined here
% rather than inline so that no new package is needed for the algorithm.
\providecommand{\algcmt}[1]{{\normalfont\itshape // #1}}
\providecommand{\algind}[1]{\hspace*{#1em}}

\appendix

\setcounter{secnumdepth}{3}

\section{DOPA Dataset Statistics and Split Details}
\label{app:dopa_details}

Table~\ref{tab:dopa_statistics} summarizes the overall statistics of DOPA. The dataset contains 5,000 images, each exhaustively annotated against a vocabulary $\mathcal{V}$ of 160 canonical object labels. To support matching free-form object mentions in model generations, the vocabulary is further associated with an alias map $\mathcal{A}$ containing 894 surface forms, with an average of 5.59 aliases per canonical label. Each image contains, on average, 10.75 positively annotated concepts.
Importantly, DOPA is constructed exclusively from the training-side image pool and is strictly separated from all evaluation splits. For the Cap. Score\(^{\dagger}\) setting in the experiments, evaluation images are independently annotated using the same DOPA vocabulary and verification protocol; these annotations are used solely for evaluation and are never used during training.

\begin{table}[h]
\centering
\caption{Overall statistics of the DOPA dataset.}
\label{tab:dopa_statistics}
\begin{tabular}{l r}
\toprule
\textbf{Statistic} & \textbf{Value} \\
\midrule
Number of images & 5,000 \\
Canonical object labels ($|\mathcal{V}|$) & 160 \\
Aliases ($|\mathcal{A}|$) & 894 \\
Average aliases per canonical label & 5.59 \\
Average positive annotations per image & 10.75 \\
\bottomrule
\end{tabular}
\end{table}

The occurrence frequencies of the canonical labels exhibit a pronounced long-tailed distribution (Table~\ref{tab:dopa_label_distribution}). Nearly half of the vocabulary (76 labels, 47.5\%) appears in fewer than 100 images. In contrast, only 14 labels (8.8\%) appear in at least 1,000 images. This distribution reflects that beyond frequently occurring objects, DOPA retains a substantial number of less frequent concepts that can nevertheless appear in model generations and therefore require reliable reward supervision.

\begin{table}[h]
\centering
\caption{Label distribution according to the number of images in which they are annotated as present.}
\label{tab:dopa_label_distribution}
\begin{tabular}{l c r}
\toprule
\textbf{Number of positive images} &
\textbf{\# Labels} & \textbf{Percentage} \\
\midrule
$<100$ & 76 & 47.5\% \\
$100$--$499$ & 59 & 36.9\% \\
$500$--$999$ & 11 & 6.9\% \\
$\geq1{,}000$ & 14 & 8.8\% \\
\midrule
Total & 160 & 100\% \\
\bottomrule
\end{tabular}
\end{table}

\section{Full Formalization of SCAPO}
\label{app:formal}

\subsection{Claim Sets and Problem Formulation}
\label{app:formulation}

This section details the formal definitions and mathematical formulations omitted from the main text for brevity. Given an image $x$ and a prompt $p$, we sample $G$ responses from the current policy snapshot:
\begin{equation}
y_i \sim \pi_{\theta_{\mathrm{old}}}(\cdot\mid x,p),
\qquad i=1,\ldots,G,
\end{equation}
and segment each response at subsentence boundaries:
\begin{equation}
y_i=[s_{i,1},s_{i,2},\ldots,s_{i,M_i}].
\end{equation}

For each subsentence $s_{i,j}$ we extract its object mentions and map them to canonical labels through the vocabulary $\mathcal{V}$ and the alias map $\mathcal{A}$. Consulting the DOPA object-existence matrix $E^\ast$ for image $x$, we partition the matched labels into the supported set and the hallucinated set:
\begin{equation}
P^{+}_{i,j} = \{\,\ell : \ell \text{ mentioned in } s_{i,j},\; E^\ast_{\ell,x}=1\,\},
\qquad
P^{-}_{i,j} = \{\,\ell : \ell \text{ mentioned in } s_{i,j},\; E^\ast_{\ell,x}=0\,\}.
\end{equation}
Because DOPA annotates every label in $\mathcal{V}$ as either present or absent for every image, this partition is total over in-vocabulary mentions: there is no third ``unknown'' category, which is exactly the ambiguity that positive-only annotation sources leave unresolved. Entities that do not map into $\mathcal{V}$ are excluded from judgment, so a subsentence whose only object mentions are out-of-vocabulary has $n^{\mathrm{new}}_{i,j}=n^{\mathrm{rep}}_{i,j}=n^{\mathrm{hal}}_{i,j}=0$ and falls into the ``no objects'' branch of Eq.~\eqref{eq:subsentence_reward}.

To reward informativeness without rewarding repetition, we track the supported objects already asserted earlier in the same response via a history set
\begin{equation}
H_{i,j} = \bigcup_{k=1}^{j-1}P^{+}_{i,k},
\qquad\text{with } H_{i,1}=\varnothing,
\end{equation}
which splits the current supported claims into novel and repeated claims:
\begin{equation}
P^{\mathrm{new}}_{i,j} = P^{+}_{i,j}\setminus H_{i,j},
\qquad
P^{\mathrm{rep}}_{i,j} = P^{+}_{i,j}\cap H_{i,j}.
\end{equation}
The counts appearing in Eq.~\eqref{eq:subsentence_reward} are then simply the
cardinalities of these sets:
\begin{equation}
n^{\mathrm{new}}_{i,j}=|P^{\mathrm{new}}_{i,j}|,
\qquad
n^{\mathrm{rep}}_{i,j}=|P^{\mathrm{rep}}_{i,j}|,
\qquad
n^{\mathrm{hal}}_{i,j}=|P^{-}_{i,j}|.
\end{equation}
Note that $H_{i,j}$ is accumulated over supported claims only and is reset for each response, so the novelty bonus is defined strictly within each sampled response. The optimization target is thus to shrink $P^{-}_{i,j}$, expand $P^{\mathrm{new}}_{i,j}$, and discourage excessive $P^{\mathrm{rep}}_{i,j}$, jointly guiding the model toward highly informative and faithful generations.

\subsection{The SCAPO Optimization Step}
\label{app:pseudocode}

\begin{figure}[t]
\centering
\small
\begin{minipage}{0.95\textwidth}
\hrule\vspace{0.4em}
\textbf{Algorithm 1}\quad One SCAPO optimization step
\vspace{0.3em}\hrule\vspace{0.5em}

\begin{flushleft}
\textbf{Input:} policy $\pi_\theta$; image--prompt batch $\mathcal{B}$; DOPA
matrix $E^\ast$; vocabulary $\mathcal{V}$; alias map $\mathcal{A}$;
coefficients $r_g,r_{\mathrm{rep}},r_h,r_{\mathrm{reg}}$; weighting functions
$\lambda_g,\lambda_h$; group size $G$; clip range $\epsilon$ \\
\textbf{Output:} updated parameters $\theta$
\end{flushleft}
\vspace{-0.3em}
\begin{tabular}{@{}r@{\hspace{1em}}l@{}}
 1 & $\pi_{\theta_{\mathrm{old}}}\leftarrow\pi_\theta$ \quad \algcmt{freeze the rollout policy} \\
 2 & $\mathcal{D}\leftarrow\varnothing$ \quad \algcmt{rollout buffer for this step} \\
 3 & \textbf{for each} $(x,p)\in\mathcal{B}$ \textbf{do} \\
 4 & \algind{1} \textbf{for} $i=1$ \textbf{to} $G$ \textbf{do} \\
 5 & \algind{2} $y_i\sim\pi_{\theta_{\mathrm{old}}}(\cdot\mid x,p)$ \\
 6 & \algind{2} $[s_{i,1},\ldots,s_{i,M_i}]\leftarrow\textsc{SplitSubsentences}(y_i)$ \\
 7 & \algind{2} $\{T_{i,j}\}\leftarrow\textsc{MapToTokenSpans}(y_i,\{s_{i,j}\})$ \quad \algcmt{spans partition the response} \\
 8 & \algind{2} $H\leftarrow\varnothing$ \quad \algcmt{supported objects seen so far in this rollout} \\
 9 & \algind{2} \textbf{for} $j=1$ \textbf{to} $M_i$ \textbf{do} \\
10 & \algind{3} $O\leftarrow\textsc{MatchLabels}(s_{i,j},\mathcal{V},\mathcal{A})$ \\
11 & \algind{3} $P^{+}\leftarrow\{\ell\in O: E^\ast_{\ell,x}=1\}$;\quad $P^{-}\leftarrow\{\ell\in O: E^\ast_{\ell,x}=0\}$ \\
12 & \algind{3} $n^{\mathrm{new}}\leftarrow|P^{+}\setminus H|$;\quad $n^{\mathrm{rep}}\leftarrow|P^{+}\cap H|$;\quad $n^{\mathrm{hal}}\leftarrow|P^{-}|$ \\
13 & \algind{3} \textbf{if} $n^{\mathrm{hal}}>0$ \textbf{then} $r_{i,j}\leftarrow-r_h\,\lambda_h(n^{\mathrm{hal}})$ \\
14 & \algind{3} \textbf{else if} $O=\varnothing$ \textbf{then} $r_{i,j}\leftarrow-r_{\mathrm{reg}}$ \\
15 & \algind{3} \textbf{else} $r_{i,j}\leftarrow r_g\,\lambda_g(n^{\mathrm{new}})+r_{\mathrm{rep}}\,\lambda_g(n^{\mathrm{rep}})$ \\
16 & \algind{3} $H\leftarrow H\cup P^{+}$ \quad \algcmt{only supported objects enter the history} \\
17 & \algind{3} $A_{i,t}\leftarrow r_{i,j}$ \textbf{for all} $t\in T_{i,j}$ \quad \algcmt{no group normalization} \\
18 & \algind{2} \textbf{end for} \\
19 & \algind{2} $\mathcal{D}\leftarrow\mathcal{D}\cup\{(x,p,y_i,\{A_{i,t}\})\}$ \\
20 & \algind{1} \textbf{end for} \\
21 & \textbf{end for} \\
22 & $N_{\mathrm{tok}}\leftarrow\sum_{(x,p,y_i,\cdot)\in\mathcal{D}}|y_i|$ \\
23 & \textbf{for} each inner update epoch \textbf{do} \\
24 & \algind{1} $\rho_{i,t}\leftarrow\pi_\theta(y_{i,t}\mid\cdot)\,/\,\pi_{\theta_{\mathrm{old}}}(y_{i,t}\mid\cdot)$ \\
25 & \algind{1} $\mathcal{L}\leftarrow-\dfrac{1}{N_{\mathrm{tok}}}\displaystyle\sum_{i,t}\min\!\big(\rho_{i,t}A_{i,t},\ \mathrm{clip}(\rho_{i,t},1-\epsilon,1+\epsilon)A_{i,t}\big)$ \\[0.6em]
26 & \algind{1} $\theta\leftarrow\theta-\eta\nabla_\theta\mathcal{L}$ \\
27 & \textbf{end for} \\
28 & \textbf{return} $\theta$ \\
\end{tabular}
\vspace{0.4em}\hrule
\end{minipage}
\caption{Pseudocode for one SCAPO optimization step.}
\label{alg:scapo}
\end{figure}

Figure~\ref{alg:scapo} outlines the complete pseudocode for a single SCAPO optimization step. Crucially, the \textsc{MatchLabels} procedure relies entirely on a deterministic lookup against the vocabulary $\mathcal{V}$ and the alias map $\mathcal{A}$, rather than invoking an external evaluation model. This design computationally decouples the rollout scoring latency from the heavy judge model used during DOPA construction. By utilizing DOPA as a reusable reward oracle, SCAPO enables highly efficient, dense per-subsentence reward assignment directly within the on-policy training loop.

\section{Implementation Details}
\label{app:impl}

\subsection{Training Configuration}
\label{app:train_config}

Table~\ref{tab:train_hparams} details the hyperparameters and configurations required to fully reproduce the SCAPO training process. To ensure a fair and rigorous comparison, all on-policy baselines (e.g., GRPO, DAPO, Dr.\ GRPO) share this identical configuration space where applicable, differing strictly in their respective advantage formulations.

\begin{table}[t]
\centering
\caption{SCAPO training configuration. Values apply to all reported SCAPO
results unless stated otherwise.}
\label{tab:train_hparams}
\small
\setlength{\tabcolsep}{6.0pt}
\begin{tabular}{ll}
\toprule
Setting & Value \\
\midrule
\multicolumn{2}{l}{\textit{Model}} \\
Backbone & Qwen2.5-VL-7B-Instruct \\
Tuning scheme & Full fine-tuning \\
Vision tower & Frozen \\
Parameter dtype & bfloat16 \\
\midrule
\multicolumn{2}{l}{\textit{Training}} \\
Optimizer & AdamW \\
Learning rate & $2\times10^{-6}$ \\
Gradient clipping & 1.0 \\
Prompts per optimization step & 128 \\
Rollouts per prompt & 8 \\
Mini-batch size & 128 (equal to the batch size, so one update per batch) \\
Micro-batch size per GPU & 2 \\
Inner update epochs per batch & 1 \\
Epochs over the training set & 8 \\
Total optimization steps & 312 ($\lfloor 5000/128 \rfloor = 39$ steps per epoch) \\
Clipping coefficient $\epsilon$ & 0.2 \\
KL loss coefficient & $0.01$ \\
\midrule
\multicolumn{2}{l}{\textit{SCAPO Reward}} \\
Novel-supported coefficient $r_g$ & 1.0 \\
Repeated-supported coefficient $r_{\mathrm{rep}}$ & 0 \\
Hallucination coefficient $r_h$ & 1.0 \\
No-object penalty $r_{\mathrm{reg}}$ & 0.1 \\
Informativeness scaling $\lambda_g(n)$ & $\min(n,1)$ \\
Hallucination scaling $\lambda_h(n)$ & $\min(n,1)$ \\
\midrule
\multicolumn{2}{l}{\textit{Environment}} \\
RL framework & verl 0.8.0.dev \\
Orchestration & Ray 2.54.0 \\
Core software & PyTorch 2.9.1, transformers 4.57.1, vLLM 0.12.0, \\
 & \quad CUDA 12.8, FlashAttention 2.8.3 \\
Text processing & spaCy 3.8.11 (\texttt{en\_core\_web\_lg}), NLTK 3.9.3 \\
\bottomrule
\end{tabular}
\end{table}

\subsection{Evaluation Metric Definitions}
\label{app:metrics}
This section provides the formal definition of the \textbf{Caption Score} metric. For each evaluation sample $i$, let $\mathcal{O}^{\mathrm{gen}}_i$ denote the set of in-vocabulary object labels mentioned in the generated description, and let $\mathcal{O}^{\mathrm{gt}}_i$ denote the set of object labels annotated as present in the corresponding image. We compute Hal.\ Rate and Cover Rate by aggregating object counts over the entire evaluation set:
\begin{equation}
\text{Hal. Rate}
=
\frac{\sum_i |\mathcal{O}^{\mathrm{gen}}_i\setminus\mathcal{O}^{\mathrm{gt}}_i|}
{\sum_i |\mathcal{O}^{\mathrm{gen}}_i|},
\qquad
\text{Cover Rate}
=
\frac{\sum_i |\mathcal{O}^{\mathrm{gen}}_i\cap\mathcal{O}^{\mathrm{gt}}_i|}
{\sum_i |\mathcal{O}^{\mathrm{gt}}_i|}.
\end{equation}
Thus, both metrics are computed using micro-averaged object counts across the full evaluation set. Hal.\ Rate measures faithfulness (lower is better), whereas Cover Rate measures informativeness (higher is better).

Reporting these two numbers separately is what allows the ``saying less'' shortcut to hide: a model can drive Hal.\ Rate toward zero simply by emitting almost no object claims, which looks like an unambiguous win if Cover Rate is not read alongside it. We therefore define the \textbf{Caption Score} as their harmonic mean:
\begin{equation}
\label{eq:cap_score}
\text{Cap. Score}
=
\mathrm{HarMean}\big(1-\text{Hal. Rate},\ \text{Cover Rate}\big)
=
\frac{2\,(1-\text{Hal. Rate})\cdot\text{Cover Rate}}
     {(1-\text{Hal. Rate})+\text{Cover Rate}},
\end{equation}
where all quantities are in $[0,1]$ and the score is reported as a percentage. Degenerate conservatism is thus penalized structurally rather than by convention. As in the main text, Cap.\ Score is computed against the evaluation dataset's original annotations, and Cap.\ Score$^{\dagger}$ against our expanded DOPA annotations.

\subsection{Description-Augmented Discriminative Inference}
\label{app:desc_aug_formal}

This section provides the formal formulation of the description-augmented inference used in Section~4.3. Given an image $x$, the SCAPO-aligned policy first generates a faithful description $d$ conditioned on the description prompt $p_{\mathrm{desc}}$:
\begin{equation}
d\sim\pi_\theta(\cdot\mid x,p_{\mathrm{desc}}).
\end{equation}

For a subsequent discriminative query $q$, the model then conditions its final answer on the original image and the generated description jointly:
\begin{equation}
a\sim\pi_\theta\big(\cdot\mid x,d,q\big),
\end{equation}
where $a$ represents the target discriminative response. Importantly, the visual input $x$ is preserved during this discriminative phase; the generated text $d$ acts strictly as an enriched auxiliary context rather than a substitute for the visual input. This design ensures that the model retains direct access to the source image to resolve queries about minor details potentially omitted from $d$. The exact prompt template integrating $x$, $d$, and $q$ is detailed in Appendix~\ref{app:prompts}. Crucially, both inference stages utilize the same policy weights $\theta$, seamlessly executing this generative-to-discriminative transfer with zero task-specific fine-tuning.

\section{Prompts}
\label{app:prompts}

\subsection{Coarse Annotation Prompt}
\label{app:prompt_coarse}

This section provides the verification prompt used to construct the initial object-existence matrix $E^{(0)}$ (see Section~3.2). It is issued once per image--label pair, with \texttt{\{object\}} bound to a canonical label, and returns a single binary verdict.

\begin{Verbatim}[fontsize=\scriptsize,frame=single,framesep=3pt]
=== SYSTEM ===
You are a strict visual existence verifier.
Return only valid JSON. Do not include markdown, explanations, reasoning, or code
fences.

=== USER ===
Inspect the image and determine whether this exact entity is clearly visible:

Entity: {object}

Rules:
- Return "supported" if the entity itself is visible enough to identify in the image.
- Small or background instances still count when they are identifiable; do not require
  the entity to be salient or large.
- If it is too tiny, blurry, heavily occluded, cut off, only implied by context, or only
  probably present, return "unsupported".
- Do not count a related object, subtype, parent class, text mention, or background
  knowledge unless the requested entity itself is visible.
- Do not count the input image/photo itself as "picture"; only count a picture, poster,
  painting, or framed image visible inside the scene.
- Clothing and accessories count only when that item itself is visible, not merely
  because a person is present.
- Scene labels such as sky, cloud, grass, mountain, beach, street, sidewalk, court, or
  island count only when that visible scene region is clear.
- If unsure, return "unsupported".

Output exactly this JSON schema:
{"verification": "supported" or "unsupported"}
\end{Verbatim}

\subsection{Label-Specific Verification Prompts}
\label{app:prompt_refine}

As discussed in Section~3.3, the initial coarse annotation can introduce systematic ambiguities for concepts with inherently vague visual boundaries. To resolve these edge cases, we replace the general-purpose prompt with tailored, label-specific instructions. These refined prompts provide rigorous operational definitions, explicit inclusion and exclusion criteria, disambiguation guidelines against semantically adjacent categories, and strict visual evidence requirements. We provide the tailored prompt for the \texttt{book} category below as a representative example.

\begin{Verbatim}[fontsize=\scriptsize,frame=single,framesep=3pt]
=== SYSTEM ===
You are a strict visual object-label verifier.
Return only valid JSON. Do not include markdown, explanations, reasoning, or code
fences.

=== USER ===
Inspect the image and decide whether the normalized label "book" is clearly visible.

Definition:
- "book" means a visible physical book-family reading, writing, or bound document
  object.
- A book can count from its cover, spine, bound page block, or open pages.

Do not count:
- A laptop, tablet, phone, television, e-reader, or other screen displaying text or an
  image of pages when no physical book-family object is visible.
- A framed picture, poster, sign, map, calendar, loose photograph, product box, card,
  envelope, or isolated sheet/stack of ordinary paper that is not identifiable as one of
  the included document types.
- A menu, restaurant bill, receipt, ticket, label, or packaging leaflet unless it is
  clearly a booklet/brochure.
- Printed book cover art or a picture of a book on another surface.
- A bookshelf, bookcase, library, desk, reader, or writer alone. The physical
  book-family object itself must be visible.

Visibility requirement:
- The object must be directly visible enough to identify a cover/spine/pages, binding,
  or folded publication. Legible text is not required.
- Books on a shelf can count when one or more actual spines or page blocks are
  recognizable; shelf-like colored rectangles alone are insufficient.
- An open book can count even when only its pages and central binding are visible.
- Small, distant, heavily blurred, or mostly hidden rectangular objects do not count
  when they could just as plausibly be boxes, screens, or loose paper.
- If unsure whether the object is a physical book-family item, mark unsupported.

Output exactly this JSON schema:
{
  "book": {"verification": "supported" or "unsupported",
           "evidence": "short visual evidence or empty string"}
}
\end{Verbatim}

For visual concepts that are highly susceptible to mutual confusion (e.g., \texttt{knife}/\texttt{fork}, \texttt{shirt}/\texttt{jacket}, and \texttt{chair}/\texttt{bench}), independent evaluations often yield inconsistent results. We address this by verifying these pairs jointly within a single discriminative prompt. By explicitly contrasting their defining features simultaneously, we enforce strict semantic boundaries and eliminate classification overlap. The joint verification prompt for \texttt{chair}/\texttt{bench} is provided below as an example.

\begin{Verbatim}[fontsize=\scriptsize,frame=single,framesep=3pt]
=== SYSTEM ===
You are a strict visual object-label verifier.
Return only valid JSON. Do not include markdown, explanations, reasoning, or code
fences.

=== USER ===
Inspect the image and decide whether these two labels are clearly visible:

1. chair 2. bench

Definitions:
- "chair" means a visible chair-like seat intended for one person, or a clearly
  separable individual seat unit.
- Count ordinary chairs, dining chairs, office chairs, folding chairs, stools, bar
  stools, armchairs, recliners, bean bag chairs, ottomans/footstools used as a seat-like
  furniture item, booster seats, visible vehicle seats, theater seats, stadium seats,
  and spectator seats.
- Count a partially covered or occupied single-person chair/armchair/recliner when
  enough structure is still visible to identify it, such as arms on the sides, a
  backrest, legs/base, frame, or a distinct one-person seat outline.
- "bench" means a visible long or shared seat intended for more than one person, usually
  without clearly separated individual chair units.
- Count park benches, wooden or metal benches, pews, bleachers, dugout benches,
  picnic-table benches, long waiting-area benches, and bench-like seats used by people
  or posed objects.

Do not count:
- For chair: do not count benches, pews, bleachers, couches, sofas, loveseats, futons,
  beds, cribs, toilets, tables, desks, counters, shelves, boxes, stools used only as
  tables/stands, or generic surfaces just because a person could sit on them.
- For bench: do not count individual chairs, stools, armchairs, recliners, couches,
  sofas, beds, cribs, toilets, tables, counters, shelves, boxes, ledges, walls,
  railings, stairs, or generic flat surfaces just because people could sit on them.
- Do not infer either label from a person merely sitting, standing, riding, or posing.
  The actual chair or bench structure must be visible.
- A vehicle, stadium, theater, restaurant, park, or dining area does not support either
  label unless the actual seat structure is visible.
- Text, logos, drawings, pictures, reflections, shadows, or printed patterns of chairs
  or benches support neither label.

Important distinction:
- Judge "chair" and "bench" independently. First look for any individual chair-like
  seats anywhere in the image, then separately look for any shared bench-like seats. The
  presence of a bench must not cause you to ignore a visible chair, and the presence of
  a chair must not cause you to ignore a visible bench.
- A single one-person seat supports "chair", not "bench".
- A single long/shared seat supports "bench", not "chair", even if only one person, toy,
  or animal is sitting on it.
- A small decorative bench, child bench, slatted two-person seat, pew, or bleacher
  supports "bench", not "chair", unless it is divided into clearly separate individual
  chair seats.
- A row of clearly separated stadium/theater/vehicle seats supports "chair"; it supports
  "bench" only if the seating is one continuous bench-like surface without individual
  seat units.
- A couch/sofa/loveseat supports the separate "couch" label, not "chair" or "bench",
  even if one or more people are sitting on it.
- A stool or ottoman can support "chair" only when it is visible as a one-person
  seat-like object; do not count a tiny unclear footrest or a stool acting only as a
  plant/table stand.
- It is possible for both labels to be supported when the image contains both individual
  chairs and a separate bench.

Visibility requirement:
- The chair or bench itself must be directly visible enough to identify, such as a seat
  surface, backrest, legs/base, arms, frame, slats, or a clearly recognizable seating
  unit.
- Small or background seats can count only when the individual-chair versus shared-bench
  structure is still identifiable.
- For background seating areas, if both separate folding/camping/stadium-style chairs
  and long benches are visible, mark both labels supported and cite evidence for each.
- If the supposed chair or bench is tiny, blurry, heavily occluded, cropped, hidden
  under a person or object, or only probably present, mark that label unsupported.
- If unsure, mark unsupported.

Output exactly this JSON schema:
{
  "chair": {"verification": "supported" or "unsupported",
            "evidence": "short visual evidence or empty string"},
  "bench": {"verification": "supported" or "unsupported",
            "evidence": "short visual evidence or empty string"}
}
\end{Verbatim}

\subsection{Description Prompt \texorpdfstring{$p_{\mathrm{desc}}$}{p\_desc}}
\label{app:prompt_desc}

We utilize two distinct description prompts across our experiments, each tailored to a specific evaluation phase.

For \textbf{training and generative evaluation}---specifically for the Cap.\ Score, Hal.\ Rate, and Cover Rate metrics reported in Table~\ref{tab:rule_based_main}---we adopt a standard, unconstrained captioning instruction. This ensures strict consistency and fair comparability with prior literature.

\begin{Verbatim}[fontsize=\scriptsize,frame=single,framesep=3pt]
Describe this image.
\end{Verbatim}

Conversely, for the \textbf{description-augmented discriminative pipeline}, the generated description serves as an enriched auxiliary context rather than a terminal evaluation target. Accordingly, we employ a refined prompt variant designed to elicit comprehensive details while explicitly constraining the model to objectively visible content:

\begin{Verbatim}[fontsize=\scriptsize,frame=single,framesep=3pt]
Describe this image objectively and in detail. Focus only on visible content.
\end{Verbatim}

\subsection{Description-Augmented Discriminative Prompt}
\label{app:prompt_descaug}

The prompt template utilized during the discriminative phase is shown below. Specifically, the generated description $d$ is inserted to serve as an explicit auxiliary context. The original image is concurrently supplied alongside the query, ensuring the model retains full access to the primary visual evidence.

\begin{Verbatim}[fontsize=\scriptsize,frame=single,framesep=3pt]
The same model previously described this image as follows:
{generated_description}

Now answer the question based on the image and the description above.
\end{Verbatim}

\section{Additional Experiments}
\label{app:additional}

\subsection{General Capability Retention}
\label{app:general}

Optimizing a policy strictly for hallucination mitigation risks compromising its broader multimodal reasoning capabilities, a phenomenon commonly referred to as the alignment tax. To verify that our fine-grained credit assignment does not degrade foundational performance, we evaluate the SCAPO-aligned policy across five comprehensive multimodal benchmarks: MME \citep{MME_2025}, MMBench \citep{MMBench_2024}, MMMU \citep{MMMU_2024}, LLaVA-Bench \citep{LLaVA_2023}, and MM-Vet \citep{MMVet_2024}.

\begin{table}[h]
\centering
\caption{Results on general multimodal benchmarks.}
\label{tab:general}
\small
\setlength{\tabcolsep}{4.5pt}
\resizebox{\textwidth}{!}{%
\begin{tabular}{lccccccc}
\toprule
& \multicolumn{2}{c}{MME} & \multicolumn{2}{c}{MMBench}
& MMMU & LLaVA-Bench & MM-Vet \\
\cmidrule(lr){2-3}\cmidrule(lr){4-5}\cmidrule(lr){6-6}\cmidrule(lr){7-7}\cmidrule(lr){8-8}
Method
& Percep.\ $\uparrow$ & Cognit.\ $\uparrow$
& Circular $\uparrow$ & Vanilla $\uparrow$
& Acc.\ $\uparrow$ & Rel.\ Score $\uparrow$ & Total $\uparrow$ \\
\midrule
Qwen2.5-VL & 1717.8 & 625.7 & \textbf{84.8} & 88.0 & \textbf{53.8} & 69.7 & 65.9 \\
\textbf{+ SCAPO (Ours)} & \textbf{1728.2} & \textbf{635.4} & 84.4 & \textbf{88.2} & \textbf{53.8} & \textbf{70.4} & \textbf{66.0} \\
\bottomrule
\end{tabular}
}
\end{table}

The results in Table~\ref{tab:general} demonstrate that SCAPO effectively avoids the alignment tax commonly associated with specialized fine-tuning. Across the evaluated benchmarks, six of the seven metrics remain stable or exhibit slight improvements following our alignment process. This confirms that structurally penalizing local hallucinations does not compromise the target model's foundational reasoning skills. Furthermore, we hypothesize that these marginal gains occur because explicitly rewarding valid visual claims positively reinforces the model's overall visual grounding, translating to subtle benefits on broader multimodal comprehension tasks.

\subsection{Ablation of SCAPO Components}
\label{app:reward_ablation}

Table~\ref{tab:reward_ablation} ablates reward design within SCAPO.
\textbf{Additive hallucination penalty} replaces the zero-tolerance branch in Eq.~\eqref{eq:subsentence_reward} with an additive term, allowing valid claims to numerically offset local hallucinations.
\textbf{Full reward for repeated claims} removes the discount on $r_{\mathrm{rep}}$, assigning redundant supported objects the same positive reward as novel ones.
\textbf{Without no-object regularization} drops the penalty ($-r_{\mathrm{reg}}$) applied to subsentences lacking verifiable objects.
All variants use the same training configuration (Appendix~\ref{app:train_config}) and DOPA annotations, differing strictly in their mathematical reward or advantage formulations.

\begin{table}[t]
\centering
\caption{Ablation of SCAPO's components.}
\label{tab:reward_ablation}
\small
\setlength{\tabcolsep}{6.0pt}
\begin{tabular}{lcc}
\toprule
Variant & Avg.\ Score $\uparrow$ & Hal.\ Rate $\downarrow$ \\
\midrule
Qwen2.5-VL & 3.26 & 39.6 \\
\midrule
Additive hallucination penalty & 3.48 & \textbf{33.3} \\
Full reward for repeated claims & 3.41 & 37.5 \\
Without no-object regularization & 3.46 & 34.4 \\
\midrule
\textbf{SCAPO (Ours)} & \textbf{3.52} & \textbf{33.3} \\
\bottomrule
\end{tabular}
\end{table}

As demonstrated in the results, the complete SCAPO formulation achieves the best overall score, while each ablation degrades at least one metric, supporting the complementary roles of these reward-design choices.

\subsection{Generalization Across Backbones}
\label{app:backbone}

To examine whether SCAPO extends beyond the primary backbone, we additionally evaluate it on Qwen3-VL \citep{Qwen3VL_2025}.

\begin{table}[t]
\centering
\caption{Results with Qwen3-VL backbone. Cap.\ Score$^{\dagger}$ is
computed against the expanded DOPA annotations.}
\label{tab:backbone}
\small
\setlength{\tabcolsep}{5.5pt}
\resizebox{0.9\textwidth}{!}{%
\begin{tabular}{lccccc}
\toprule
& \multicolumn{2}{c}{AMBER} & \multicolumn{2}{c}{MS COCO} & PhD \\
\cmidrule(lr){2-3}\cmidrule(lr){4-5}\cmidrule(lr){6-6}
Method
& Cap.\ Score $\uparrow$ & Dis.\ Acc.\ $\uparrow$
& Cap.\ Score $\uparrow$ & Cap.\ Score$^{\dagger}$ $\uparrow$
& PhD-all $\uparrow$ \\
\midrule
Qwen3-VL & 76.0 & 86.6 & 72.8 & 67.1 & 75.0 \\
\textbf{+ SCAPO (Ours)} & \textbf{78.8} & \textbf{87.6} & \textbf{76.2} & \textbf{73.1} & \textbf{75.8} \\
\bottomrule
\end{tabular}
}
\end{table}

As detailed in Table~\ref{tab:backbone}, the empirical results confirm the robust transferability of our approach. Specifically, SCAPO consistently enhances the performance of the Qwen3-VL backbone, demonstrating that the gains of SCAPO transfer to an additional LVLM backbone.

\section{Case Study}
\label{app:case_study}

\begin{figure}[t]
    \centering
    \includegraphics[width=\textwidth]{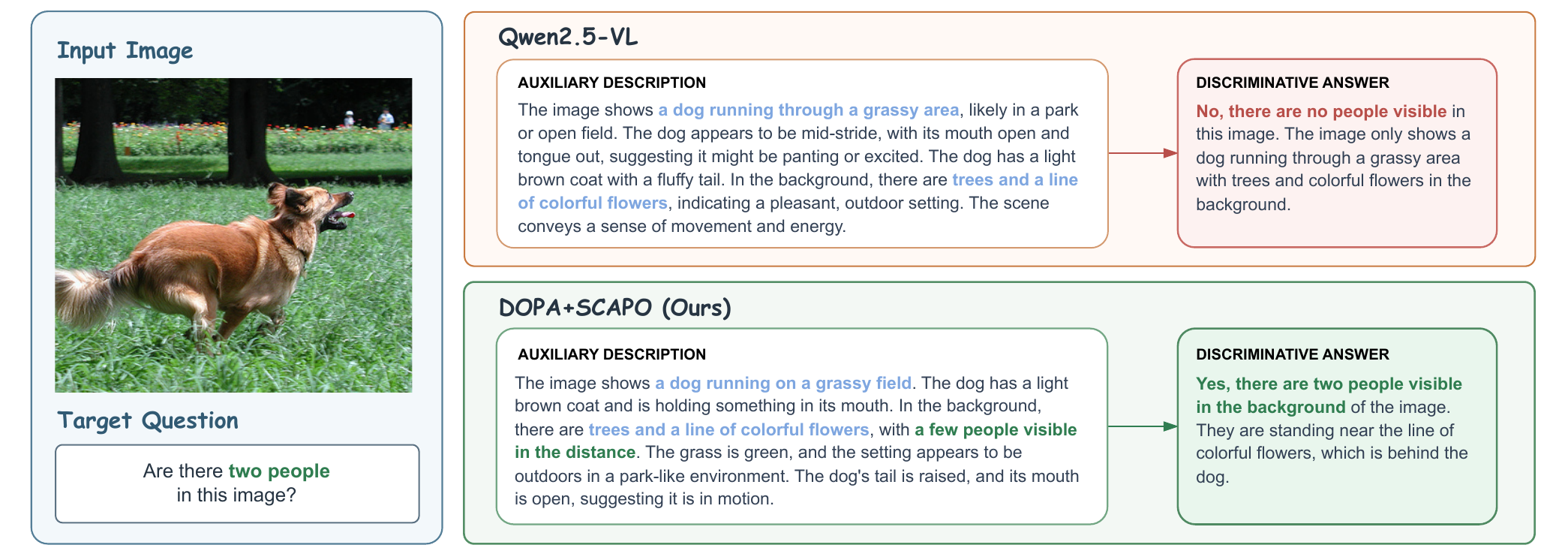}
    \caption{Qualitative comparison of description-augmented discriminative inference between the baseline (Qwen2.5-VL) and our SCAPO-aligned model.}
    \label{fig:case_study}
\end{figure}

Figure~\ref{fig:case_study} contrasts the performance of the base Qwen2.5-VL model with our SCAPO-aligned model on a representative case. Given the target question regarding small-scale background individuals, the auxiliary description generated by Qwen2.5-VL misses these subtle entities, focusing solely on the dog and general scenery. Consequently, the missing visual evidence leaves the subsequent discriminative prediction without sufficient context, resulting in a false negative. In contrast, our model produces an auxiliary description that is both highly faithful and rich in detail, successfully capturing the distant background people. This comprehensive visual grounding provides the necessary context for the model to correctly resolve the query.

\section{Limitations}
\label{app:limitations}

Our framework currently operates over a predefined object vocabulary, which leaves two practical directions for further extension. Although DOPA provides substantially denser supervision than conventional annotations, concepts outside $\mathcal{V}$ are still not explicitly evaluated and therefore receive limited training signal. This coverage can be progressively improved by expanding the vocabulary with newly observed model rollouts. In addition, constructing dense presence--absence annotations introduces extra offline annotation and verification cost, particularly when extending to larger vocabularies or new visual domains. This overhead is confined to dataset construction rather than inference, and future work could further reduce it through model-assisted annotation and iterative vocabulary expansion.

\end{document}